\documentclass{llncs}
\usepackage{llncsdoc}

\usepackage{tikz}
\usetikzlibrary{trees,snakes}

\usepackage{url}

\usepackage[ruled,vlined,linesnumbered]{algorithm2e}
\usepackage{algorithmic}

\usepackage{caption}
\usepackage[]{subfig}
\usepackage{float}

\usepackage{breakcites}

\usepackage{array}
\usepackage{amssymb}
\usepackage{amsmath}

\usepackage{array,multirow,graphicx}

\usepackage{textcomp}
\usepackage{scalefnt}
\usepackage{color, colortbl}
\definecolor{orange}{rgb}{1,0.5,0}
\newcommand{\comment}[2]{\hspace{-0.0001px}{\scalefont{#1}\textcolor{gray}{\textit{#2}}}}

\makeatletter
\def\grd@save@target#1{%
  \def\grd@target{#1}}
\def\grd@save@start#1{%
  \def\grd@start{#1}}
\tikzset{
  grid with coordinates/.style={
    to path={%
      \pgfextra{%
        \edef\grd@@target{(\tikztotarget)}%
        \tikz@scan@one@point\grd@save@target\grd@@target\relax
        \edef\grd@@start{(\tikztostart)}%
        \tikz@scan@one@point\grd@save@start\grd@@start\relax
        \draw[minor help lines] (\tikztostart) grid (\tikztotarget);
        \draw[major help lines] (\tikztostart) grid (\tikztotarget);
        \grd@start
        \pgfmathsetmacro{\grd@xa}{\the\pgf@x/1cm}
        \pgfmathsetmacro{\grd@ya}{\the\pgf@y/1cm}
        \grd@target
        \pgfmathsetmacro{\grd@xb}{\the\pgf@x/1cm}
        \pgfmathsetmacro{\grd@yb}{\the\pgf@y/1cm}
        \pgfmathsetmacro{\grd@xc}{\grd@xa + \pgfkeysvalueof{/tikz/grid with coordinates/major step}}
        \pgfmathsetmacro{\grd@yc}{\grd@ya + \pgfkeysvalueof{/tikz/grid with coordinates/major step}}
        \foreach \x in {\grd@xa,\grd@xc,...,\grd@xb}
        \node[anchor=north] at (\x,\grd@ya) {\pgfmathprintnumber{\x}};
        \foreach \y in {\grd@ya,\grd@yc,...,\grd@yb}
        \node[anchor=east] at (\grd@xa,\y) {\pgfmathprintnumber{\y}};
      }
    }
  },
  minor help lines/.style={
    help lines,
    step=\pgfkeysvalueof{/tikz/grid with coordinates/minor step}
  },
  major help lines/.style={
    help lines,
    line width=\pgfkeysvalueof{/tikz/grid with coordinates/major line width},
    step=\pgfkeysvalueof{/tikz/grid with coordinates/major step}
  },
  grid with coordinates/.cd,
  minor step/.initial=.2,
  major step/.initial=1,
  major line width/.initial=2pt,
}
\makeatother

\begin{document}

\title{Rapid Building Detection using Machine Learning}

\author{%
Joseph Paul Cohen\inst{1} \and %
Wei Ding\inst{1} \and%
Caitlin Kuhlman\inst{1} \and %
\newline Aijun Chen\inst{2}  \and%
Liping Di\inst{2} %
}


\institute{
Department of Computer Science, University of Massachusetts Boston, \{joecohen,ckuhlman,ding\}@cs.umb.edu
\and
Spatial Information Science and Systems, College of Science, George Mason University, \{achen6,ldi\}@gmu.edu
}

\maketitle
\begin{abstract}
This work describes algorithms for performing discrete object detection, specifically in the case of buildings, where usually only low quality RGB-only geospatial reflective imagery is available. We utilize new candidate search and feature extraction techniques to reduce the problem to a machine learning (ML) classification task. Here we can harness the complex patterns of contrast features contained in training data to establish a model of buildings. We avoid costly sliding windows to generate candidates; instead we innovatively stitch together well known image processing techniques to produce candidates for building detection that cover 80-85\% of buildings. Reducing the number of possible candidates is important due to the scale of the problem. Each candidate is subjected to classification which, although linear, costs time and prohibits large scale evaluation. We propose a candidate alignment algorithm to boost classification performance to 80-90\% precision with a linear time algorithm and show it has negligible cost. Also, we propose a new concept called a Permutable Haar Mesh (PHM) which we use to form and traverse a search space to recover candidate buildings which were lost in the initial preprocessing phase. All code and datasets from this paper are made available online\footnote{http://kdl.cs.umb.edu/w/datasets/}.
\end{abstract}

\section{Introduction}

Rapid detection and classification of discrete objects such as buildings in geospatial imagery has many applications such as damage assessments by comparing before and after building detections \cite{voigt_satellite_2007} \cite{dong_comprehensive_2013} \cite{brunner_earthquake_2010}. Large scale change detection at an object level can enable computer assisted updating of maps by identifying new or removed objects between multiyear satellite imagery \cite{bonnefon_geographic_2002}. This could also allow for the next evolution of the USGS National Land Cover Database (NLCD) analysis \cite{xian_change_2011}. Also, in a national security interest and in the funding motivation of this research, ontological analysis can be performed using the spatial arrangement of groups of buildings to identify large manufacturing, power generation, and weapons proliferation sites.

Problems restrict the usage of existing research which require camera alignment information (azimuth and zenith angles) and/or special equipment that captures near-infrared channels. Runtime is also a large factor which restricts the scale of deployment. In this work we present a combination of methods which have minimum imagery requirements (they work on common grayscale imagery) and provides scale and rotation invariant detection with a relatively inexpensive computation. 

The first contribution of this paper is our method does not depend on sliding windows to generate building candidates (Section \ref{section:candidates:method}). Building candidates are rectangles, identified by a center, height, width, and rotation, that likely contain a building. If these were generated using a brute force sliding window approach processing an image very expensive because the centers can be any pixel, the width and height can be any combination (non-overlapping), and the rotation can be between $0^\circ-180^\circ$. We devise a linear time strategy utilizing building shadows as a major feature because they are high contrast  straight `L' shaped feature unique to man made objects \cite{irvin_methods_1989} \cite{lin_building_1998} and \cite{karantzalos_recognition-driven_2009}.

The second contribution is how we align buildings in linear time to increase classification accuracy (Section \ref{section:rotate}). We utilize a summation of Gaussians each centered and scaled depending on the direction and magnitude of the vectors that form the contour around a building. We describe a linear time algorithm for computing this and show it has a negligible cost as well as a significant performance gain of up to 5\% accuracy. 

The third contribution is our candidate Permutable Haar Mesh (PHM) search method that heuristically searches nearby candidate boxes to find buildings via a greedy graph search algorithm (Section \ref{section:phm}). Because we utilize Haar contrast features \cite{viola_robust_2004} for their supreme performance; if our building candidate box does not properly cover the building it will not be considered a building because its feature distributions won't align to learned examples. The PHM approach is expensive and is not part of our rapid solution but can be employed to increase accuracy if it is really necessary.

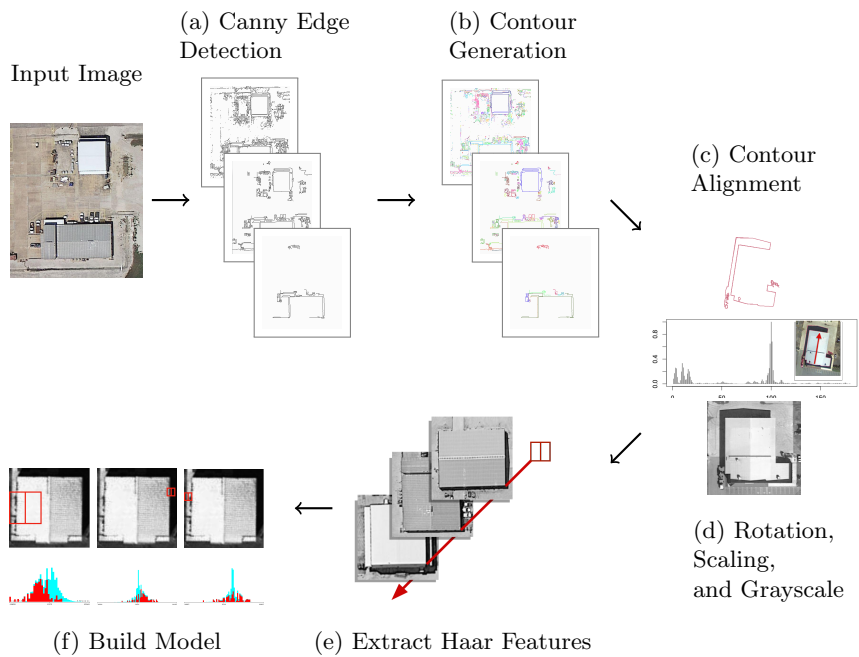
\begin{figure}[h]
\begin{tikzpicture}


\node[inner sep=0pt] (rtm) at (-1,3.6){\includegraphics[width=50pt]
    {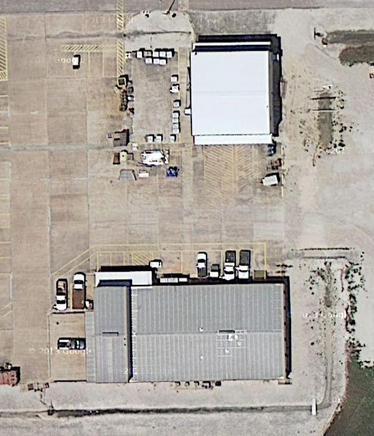}};

\draw[thick,style={sloped,anchor=north,auto=false}] (-1,5) 
 	node[above] {Input Image} (1,5);

\draw[thick,->,style={sloped,anchor=north,auto=false}] (0,3.6) -- 
 	node[above] {} (0.5,3.6);

\node at (1.5,5.8) [align=left]{(a) Canny Edge\\Detection};

\node[inner sep=0pt] (rtm) at (1.3,4.5)
    {\fcolorbox{gray}{white}{\includegraphics[width=30pt]
    {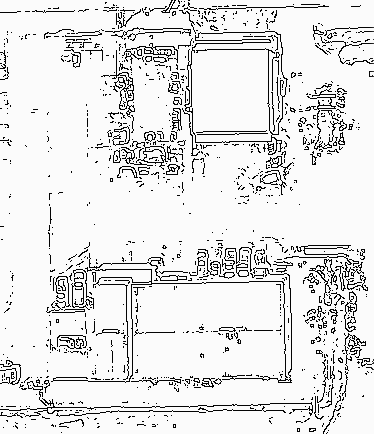}}};
    
\node[inner sep=0pt] (rtm) at (1.6,3.5)
    {\fcolorbox{gray}{white}{\includegraphics[width=30pt]
    {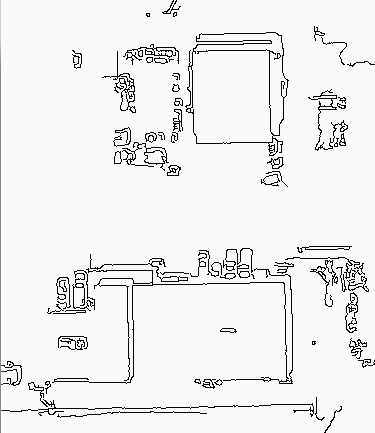}}};
    
\node[inner sep=0pt] (rtm) at (2.0,2.5)
    {\fcolorbox{gray}{white}{\includegraphics[width=30pt]
    {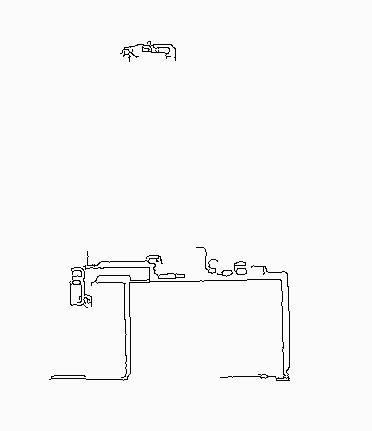}}};

\draw[thick,->,style={sloped,anchor=north,auto=false}] (3,3.6) -- 
 	node[above] {} (3.5,3.6);

\node at (4.8,5.8) [align=left]{(b) Contour\\Generation};

\node[inner sep=0pt] (rtm) at (4.5,4.5)
    {\fcolorbox{gray}{white}{\includegraphics[width=30pt]
    {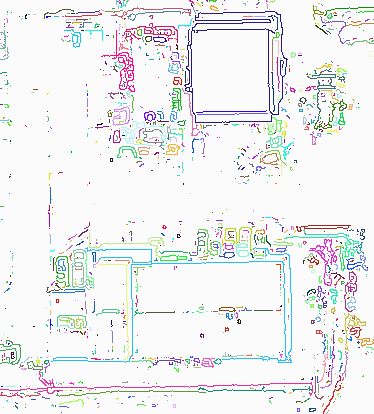}}};
    
\node[inner sep=0pt] (rtm) at (4.9,3.5)
    {\fcolorbox{gray}{white}{\includegraphics[width=30pt]
    {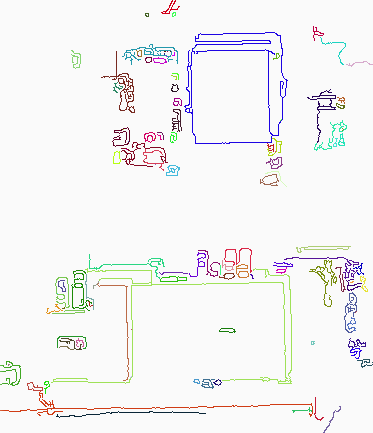}}};
    
\node[inner sep=0pt] (rtm) at (5.3,2.5)
    {\fcolorbox{gray}{white}{\includegraphics[width=30pt]
    {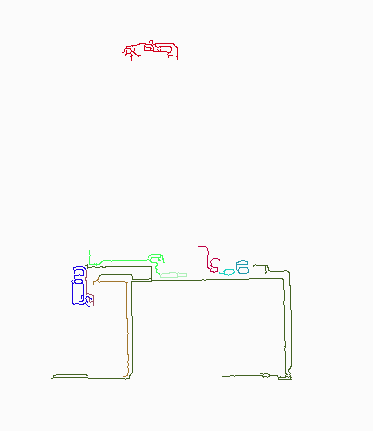}}};

\draw[thick,->,style={sloped,anchor=north,auto=false}] (6.1,3.6) -- 
 	node[above] {} (6.5,3.2);

\node at (8,4)  [align=left]{(c) Contour\\Alignment};

\node[inner sep=0pt] (rtm) at (8,2.6)                      
    {\includegraphics[height=35pt]{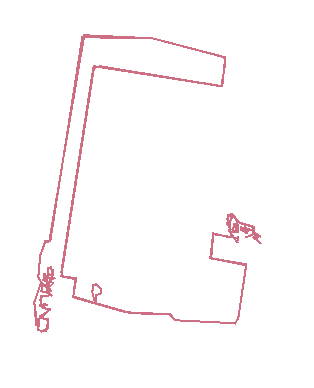}};
\node[inner sep=0pt] (rtm) at (8,1.5)    
    {\includegraphics[trim = 12mm 18mm 0mm 0mm, clip,height=30pt]{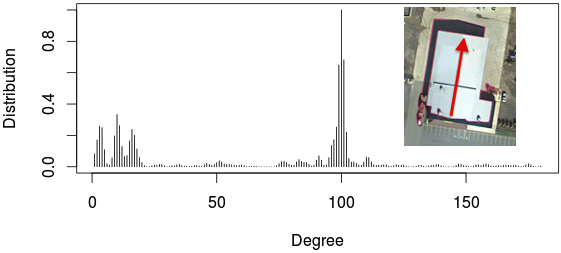}};
\node[inner sep=0pt] (rtm) at (8,0.3)  
    {\includegraphics[height=35pt]{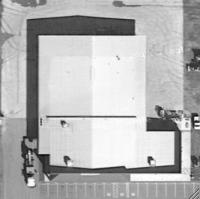}};

\draw[thick,->,style={sloped,anchor=north,auto=false}] (6.5,0.5) -- 
 	node[above] {} (6.1,0.1);

\node at (8.2,-1.2)  [align=left]{(d) Rotation,\\Scaling,\\and Grayscale};
    
\node[inner sep=0pt] (rtm) at (4,-0.5)
    {\includegraphics[height=70pt]{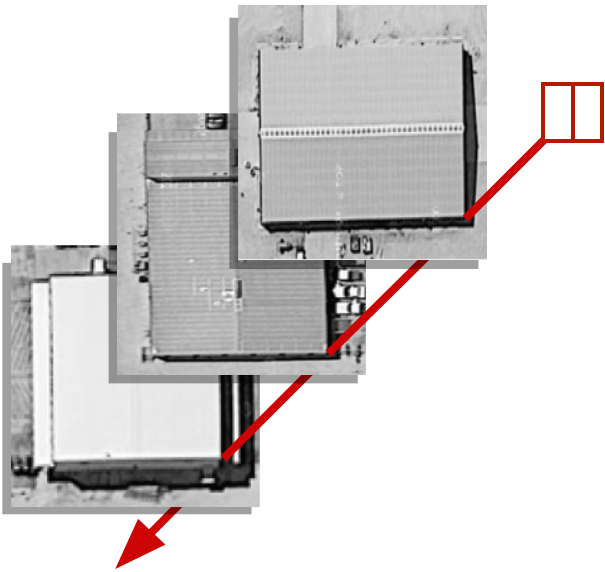}};

\node at (4,-2.3) [align=left]{(e) Extract Haar Features};

\node[inner sep=0pt] (rtm) at (-0.2,-0.5){  
  \includegraphics[width=30pt]{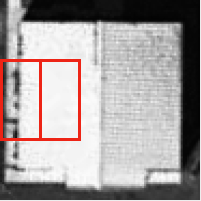}
  \includegraphics[width=30pt]{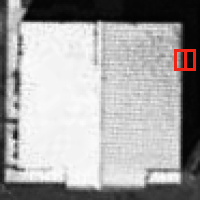}
  \includegraphics[width=30pt]{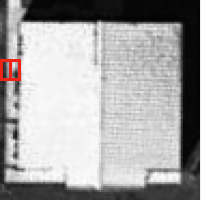}
};
\node[inner sep=0pt] (rtm) at (-0.2,-1.5){
  \includegraphics[width=30pt]{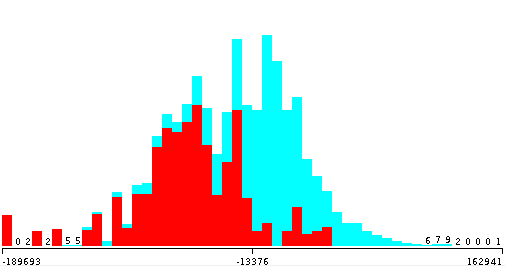}
  \includegraphics[width=30pt]{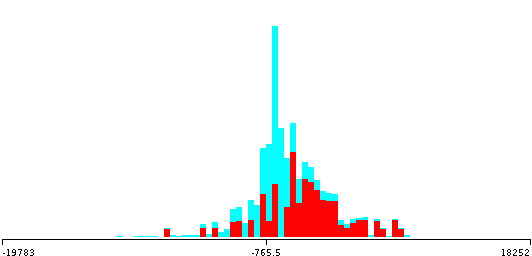}
  \includegraphics[width=30pt]{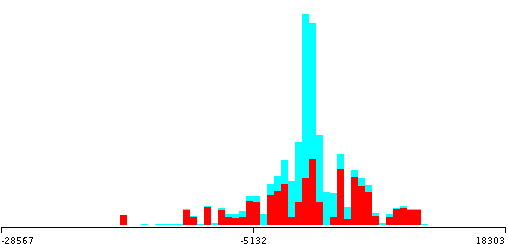}
  };

\draw[thick,->,style={sloped,anchor=north,auto=false}] (2.4,-0.5) -- 
 	node[above] {} (1.9,-0.5);

\node at (-0.2,-2.3) [align=left]{(f) Build Model};

\end{tikzpicture}
\caption{
An overview of the method is shown. First (a) Canny edge detection is run using a range of threshold values. The resulting binary images are processed for contours (b). Each contour is considered a candidate. These contours have their alignment detected (Note: a different building is used to illustrate this) (c). They are then rotated to standard alignment, scaled to a standard size, and converted to grayscale (d). For every candidate, Haar image masks are extracted from fixed locations to capture contrast (e). Next these contrast values are discriminative enough to build a model and make accurate predictions (f).
}
\label{fig:overview}
\end{figure}

\section{Method}
\label{section:method}

An overview of our method is shown in Figure \ref{fig:overview}. First (in Figure \ref{fig:overview}a) Canny edge detection is run using a range of threshold values. The Canny edge detection \cite{canny_computational_1986} a fast straightforward method uses high and low thresholds to determine edges and using only one set of threshold values would not discover all buildings (Discussed in Section \ref{section:candidates:method}). Instead, all possible combinations of threshold values are used limited by a step size between the values. The resulting binary images are processed for contours (Figure \ref{fig:overview}b) in linear time \cite{chang_linear-time_2004}. Each contour is considered a candidate. Some of the resulting contours are filtered out based on a minimum number of pixels that can be used for prediction and if they are redundant to other contours by only differing by less than 5 pixels.

These contours have their alignment (Figure \ref{fig:overview}c) detected automatically (Note: in the figure a different building is used to illustrate this). Section \ref{section:rotate} discusses the rotation method. The candidates are then automatically rotated to a standard alignment, scaled to a standard size, and converted to grayscale for Haar feature extraction (Figure \ref{fig:overview}d). This rotation is so the Haar features will have more correlation when a model is built.

For every candidate, Haar features are extracted from fixed locations to capture contrast (Figure \ref{fig:overview}e). Haar features have been successful and proved rapid and robust by \cite{viola_robust_2004}. To extract a Haar feature, a rectangle is first overlaid at a specific and consistent location on the image. The rectangle is split in half and the pixels inside each half are summed and subtracted from each other. The resulting value represents the contrast at that location in the image and can be compared to other images. Combinations of these features will be discriminative enough to build a model (Figure \ref{fig:overview}f). This model can then be used to predictions when given unseen Haar feature values from a new test image.

To complement this method we present an optional step (due to computational cost) which is a novel candidate permutation method called a Permutable Haar Mesh (PHM) to increase recall of candidates via greedy graph search (Section \ref{section:phm}). Recall is an evaluation metric representing how many buildings have not been missed, this metric is complementary to precision which represents how correct each prediction is. Candidates are surrounded by a bounding box and permuted by moving their top, bottom, left, and right boundaries in order to properly cover a candidate and capture buildings that would otherwise have been missed because the candidate didn't properly cover the building.

\subsection{Candidate Generation}
\label{section:candidates:method}

We utilize building shadows as a major identifier of buildings because they are a high contrast feature which provides largely straight `L' shaped contours unique to manmade objects \cite{irvin_methods_1989} \cite{lin_building_1998} and  \cite{karantzalos_recognition-driven_2009}. Canny edge detection \cite{canny_computational_1986} is still the state of the art edge detection method that can capture these shadows well. The result of Canny edge detection is a binary image representing the edges of the input. Candidates are isolated by applying a linear time contour generation algorithm \cite{chang_linear-time_2004} which groups together edge pixels and returns a set of vectors that trace along these edges forming a contour. Each contour is considered to be a candidate building, we will also call the derived forms of this contour a candidate such as a bounding box around the contour and the image pixels within this bounding box.

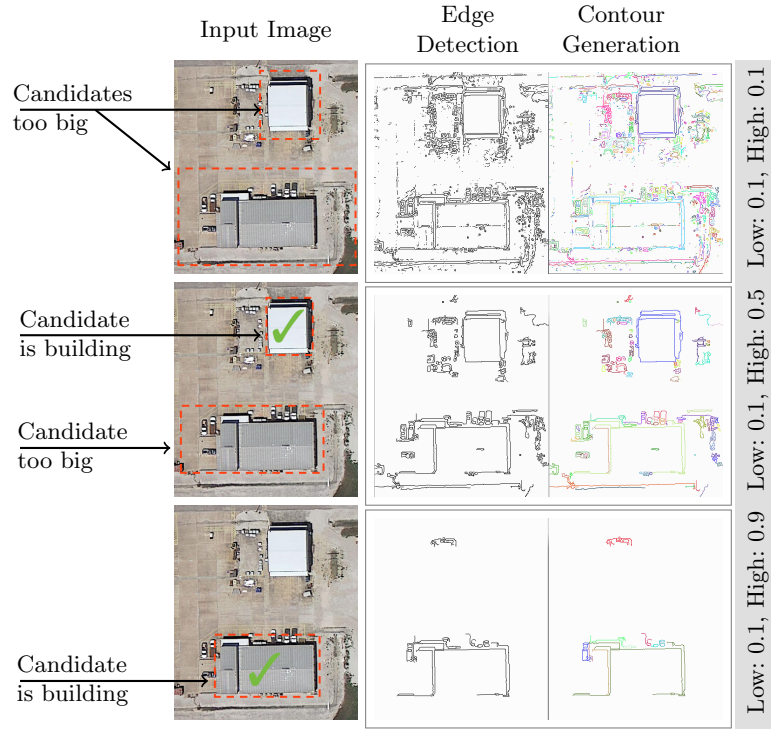
\begin{figure}[h]
\begin{center}

\newcommand{\specialcell}[2][c]{%
  \begin{tabular}[#1]{@{}c@{}}#2\end{tabular}}

\begin{tikzpicture}

\node[inner sep=0pt] (rtm) at (0,0.4){
\begin{tabular}{c c l}
Input Image & \specialcell{Edge\\ Detection} \hspace{10pt} \specialcell{Contour\\ Generation} &\\
\includegraphics[width=0.20\columnwidth]{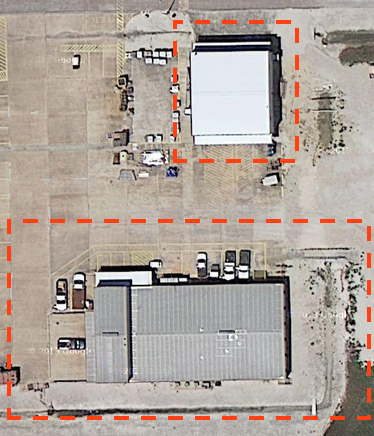} & 	\fcolorbox{gray}{white}{\includegraphics[width=0.38\columnwidth]{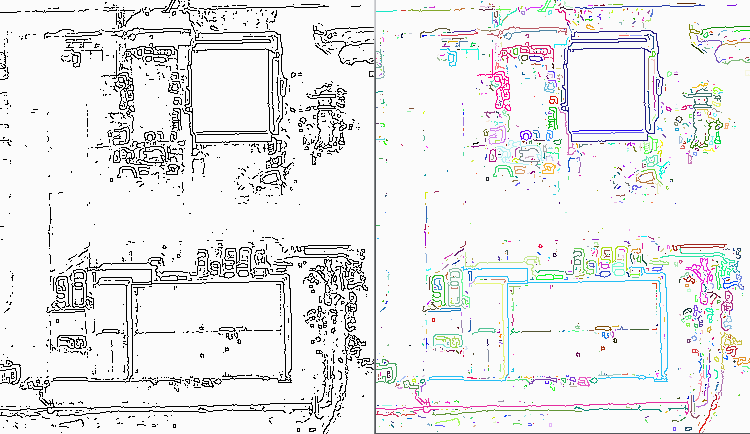}}
& 
\cellcolor{gray!25}
\rotatebox[origin=l]{90}{Low: 0.1, High: 0.1}
\\
\includegraphics[width=0.20\columnwidth]{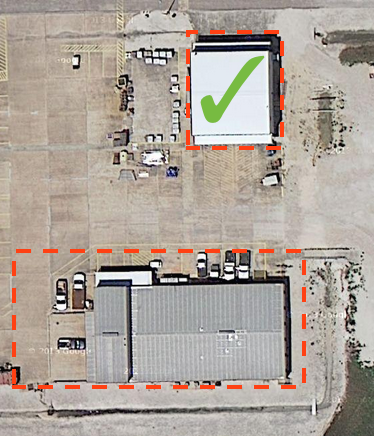} & 	\fcolorbox{gray}{white}{\includegraphics[width=0.38\columnwidth]{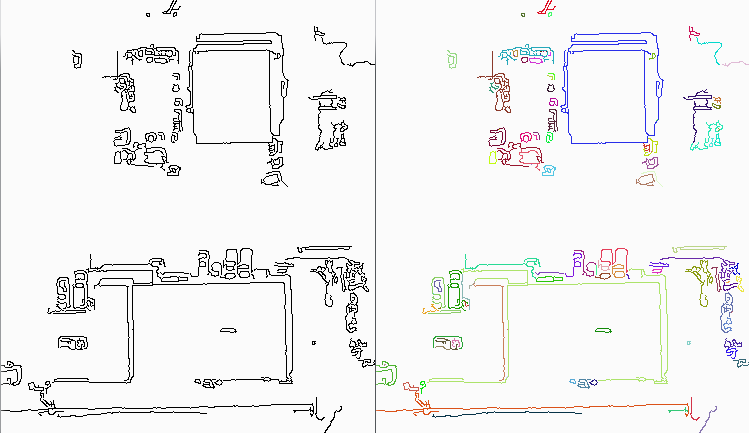}}
& 
\cellcolor{gray!25}
\rotatebox[origin=l]{90}{Low: 0.1, High: 0.5}
\\
\includegraphics[width=0.20\columnwidth]{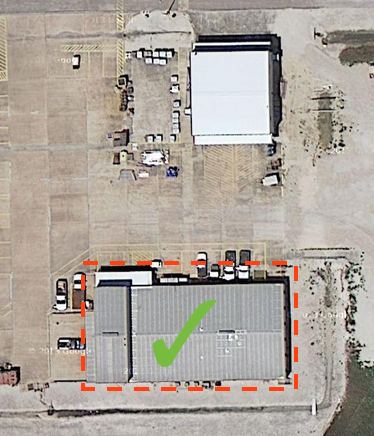} & 	\fcolorbox{gray}{white}{\includegraphics[width=0.38\columnwidth]{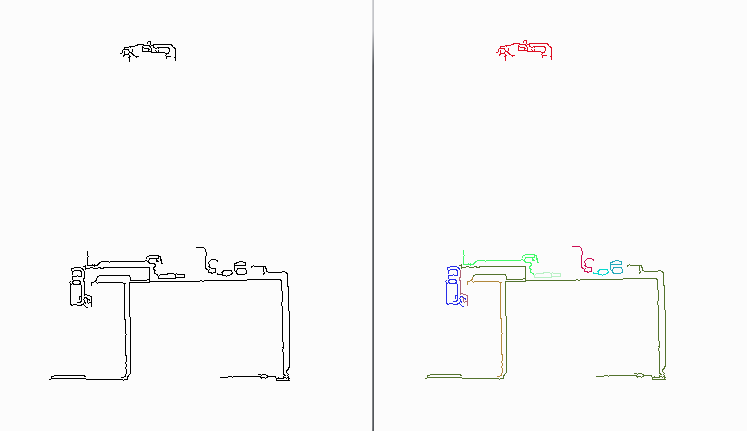}}
& 
\cellcolor{gray!25}
\rotatebox[origin=l]{90}{Low: 0.1, High: 0.9}
\end{tabular}
};
\draw[thick,->,style={sloped,anchor=north,auto=false}] (-6,0.8) -- 
 	node[above,align=left,anchor=east] {Candidate\\is building} (-2.8,0.8);
 	
\draw[thick,->,style={sloped,anchor=north,auto=false}] (-6,-3.8) -- 
 	node[above,align=left,anchor=east] {} (-3.5,-3.8);
\node at (-5.3,-3.8) [align=left]{Candidate\\is building};
 	
\draw[thick,->,style={sloped,anchor=north,auto=false}] (-6,-0.7) -- 
 	node[align=left,anchor=east] {} (-4,-0.7);
\node at (-5.3,-0.7) [align=left]{Candidate\\too big};
 	
\draw[thick,->,style={sloped,anchor=north,auto=false}] (-6,3.8) -- 
 	node[above,align=left,anchor=east] {} (-2.8,3.8);
\draw[thick,->,style={sloped,anchor=north,auto=false}] (-5,3.8) -- 
 	node[align=left,anchor=east] {} (-4,3);
\node at (-5.3,3.8) [align=left]{Candidates\\too big};
 	
\end{tikzpicture}

\caption{This figure shows the application of Canny edge detection (center) and contour detection (right) at various threshold values to generate candidates. Red dashed boxes on the left show candidates that enclose buildings and green check marks are candidates that will be classified as buildings. As the high threshold parameter to the Canny edge detector is varied from 0.1 at the top to 0.9 at the bottom different contours are generated. There is no perfect parameters to generate correct candidates for both buildings.}
\label{fig:cannythreshimage}
\end{center}
\end{figure}

Canny edge detection has two hyperparameters, a high and low thresholds for hysteresis thresholding. Canny edge detection works by first computing the gradient of the image using the contrast between pixels (scaled between 0 and 1). Gradients below the low threshold are filtered out and will not be considered edges. Gradients above the high threshold are set as edges and any remaining gradients that are connected to edges are set as edges. One combination of parameters will likely not return correct candidates for all buildings in an image as shown in Figure \ref{fig:cannythreshimage} because too high of a threshold can cause gradients of objects that neighbor buildings to become part of its contour while too low of a threshold may cause the gradients of a building not to be considered. These issues are almost always the case when buildings vary in size in the same image because gaps in high gradients along the side of a building require lower thresholds which will cause smaller buildings to be connected to neighboring objects.

In order to be scale invariant the union of the resulting contours from many different combinations of Canny threshold parameters are used to form the set of candidates. If the candidates generated in Figure \ref{fig:cannythreshimage} from the three different pairs of threshold values are merged together then all buildings will be included in the candidate set. However, as more threshold values are included, more non-buildings are included as well and create a challenge to later steps. Threshold values are chosen from a grid which is parametrized by a step size which controls the density of the grid. As the step size is decreased, more threshold values are included which results in more candidates. Section \ref{section:candidates:eval} studies the trade-off when decreasing the step size in order to maximize precision and recall.

\subsection{Building Contour Alignment}
\label{section:rotate}

Contours resulting from Chang's contour detection \cite{chang_linear-time_2004} are represented by a set of vectors $c$ and each component vector $c_i$. From these vectors we want to determine the aggregate direction of the object they represent. By rotating these candidates into alignment before extraction of the Haar features, the features become more discriminative and will result in an increase in accuracy of the trained classifier (explained in \S \ref{section:features}).

\begin{figure}[!ht]
\begin{center}

\begin{tikzpicture}[scale=1]


\node[inner sep=0pt] (rtm) at (3,7.5){\includegraphics[width=90pt] {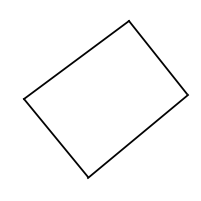}};

\node at (1,8.2)[align=right]{$\angle c_1 = 30^\circ,|c_1| = 5$};

\node at (5.3,8.5)[align=left]{$\angle c_2 = 120^\circ,|c_2| = 3$};

\node at (0.8,6.6)[align=right]{$\angle c_3 = 120^\circ,|c_3| = 3$};

\node at (5,6.7)[align=left]{$\angle c_4 = 31^\circ,|c_4| = 5$};

\draw[thick,->,style={}] (2.0,6) -- node[left,shift={(-0.5,0)}] {Using Eq \ref{eq:rot:sum}} (0,5);

\draw[thick,->,style={}] (3.6,6) -- node[right,shift={(0.5,0)}] {Using Eq \ref{eq:rot:gaus}} (5,5);

\node[inner sep=0pt] (rtm) at (-0.7,3.5){\includegraphics[width=150pt] {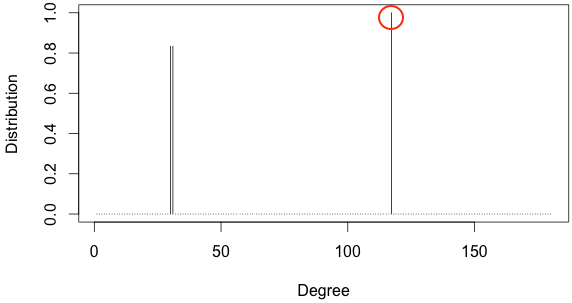}};

\node[inner sep=0pt] (rtm) at (5.3,3.5){\includegraphics[width=150pt] {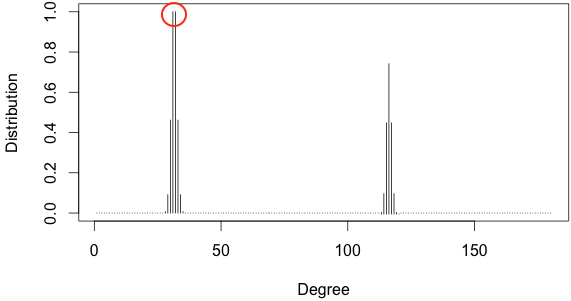}};

\draw[thick,->,style={}] (-1,4.5) -- node[shift={(0,0)},align=center] {Max\\of $120^\circ$} (0,4.5);

\draw[thick,<-,style={}] (4.5,4.5) -- node[shift={(0,0)},align=center] {Max\\of $30^\circ$} (5.5,4.5);

\end{tikzpicture}

\caption{Comparing the direction information obtained from the two discussed equations we can see a disagreement. The input contour contains four vectors ($c_i, 1\leq i \leq 4$)  Eq. \ref{eq:rot:sum} results in an aggregate angle of $120^\circ$ while Eq. \ref{eq:rot:gaus} results in the a more expected direction of $30.5^\circ$ because it is the mean of the angles. For our application rounding to $30^\circ$ and $31^\circ$ would both yield satisfactory features.}
\label{fig:contourexample}
\end{center}
\end{figure}
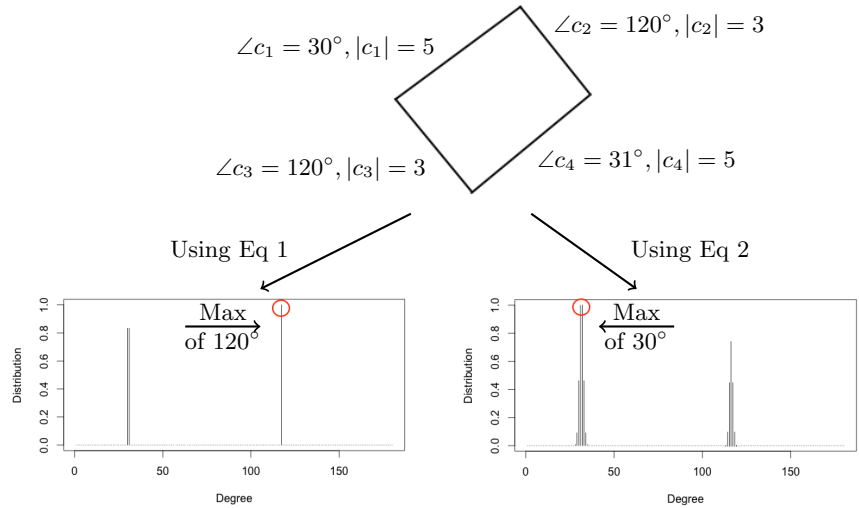

\begin{figure}[!ht]
\begin{center}
\begin{tikzpicture}[scale=1]
\node[inner sep=0pt] at (0,0){\includegraphics[width=0.49\textwidth] {AR1-bld4-can-plot.png}};
\draw[thick,->,style={}] (-0.5,1) -- node[shift={(0,0)},align=center] {Max\\of $99^\circ$} (0.5,1);
\end{tikzpicture}
\begin{tikzpicture}[scale=1]
\node[inner sep=0pt] at (0,0){\includegraphics[width=0.49\textwidth] {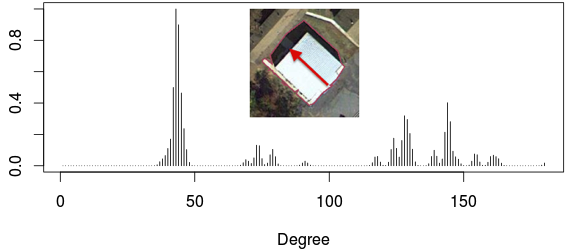}};
\draw[thick,->,style={}] (-2.3,1) -- node[shift={(0,0)},align=center] {Max\\of $42^\circ$} (-1.2,1);
\end{tikzpicture}

\caption{Histograms of the Gaussian summations of contour components evaluated at specific angles when Algorithm \ref{alg:rotatebuildings} is applied to candidates A (left) and B (right). Our method correctly identified Candidate A at a 99 degree angle and candidate B at a 42 degree angle.}
\label{fig:orentationhists}
\end{center}
\end{figure}

Determining the aggregate direction is difficult because buildings may not have their walls parallel to each other and the edge and contour detection methods may have introduced noise in the vector directions. Consider the simple example in Figure \ref{fig:contourexample}; suppose we have a contour made up of four vectors with the following directions and magnitudes $(30^\circ, 5)$, $(31^\circ, 5)$, $(120^\circ, 3)$, $(120^\circ, 3)$ which would appear to be a rectangle with the longest side as the dominant edge. If the assumption is made that the majority of the walls length will point in the dominant direction of the building then we should be able to sum the vectors with the same angle to determine which angle the majority of the walls align with. However, taking the sum for each direction would not capture the similarity of $angle(c_i) = 30^\circ$ and $31^\circ$. They would be considered independent and Eq. \ref{eq:rot:sum} would result in $120^\circ$ as the dominant direction of the building which is false. 

\begin{equation}
argmax_\theta
\displaystyle\sum_{c_i \in c}
	\{|c_i| : \theta = angle(c_i)\} 
\label{eq:rot:sum}
\end{equation}
We need to tolerate this noisy data and take these situations into account because contours can be even more complex and misleading as seen in Figure \ref{fig:orentationhists}. To accomplish this we use a method similar to a kernel density estimation which utilizes a sum of Gaussian distributions, one for each vector's degree normalized by its magnitude, shown in Equation \ref{eq:rot:gaus}.
\begin{equation}
argmax_\theta
\displaystyle\sum_{c_i \in c}
\left(
	\underbrace{\left(\frac{|c_i|}{
	\sum_{c_i \in c} |c_i|}\right)}_{\text{Normalized
	Magnitude}}
	\underbrace{\frac{1}{\sigma\sqrt{2\pi}}
	e^{-\frac{1}{2}
	\left(
		\frac{\theta - angle(c_i)}{\sigma}
	\right)^2 }}_{\text{Gaussian on orientation}}
\right)
\label{eq:rot:gaus}
\end{equation}
To determine the alignment direction we evaluate the summation for a specific input degree from $0^\circ-180^\circ$. Algorithm \ref{alg:rotatebuildings} formalizes this method.  For each contour segment $c_i$ the angle $\alpha$ is determined using the arctangent. The Gaussians are normalized based on their magnitude to the sum of all magnitudes. The maximum $\theta$ is then found by iterating over 180 possible angles. Figure \ref{fig:orentationhists} shows this method not only handles the specific issue we discussed of non parallel walls but also tolerates noise in the contour data.  Noise meaning jitter in the angle of the vectors as they wrap around the building. This can be due to pixelation error during capturing the image, contours containing vectors that don't overlap the building walls, or non-rectangular building shapes. This rotation method not only increases classification accuracy but does so with negligible increase in time (shown in \S ref{section:lineareval}).

\begin{algorithm}[!h]

\caption{Rotate Building Candidate}
\label{alg:rotatebuildings}
\KwIn{Contour $c$}
\KwOut{Rotated Contour $c_{rot}$}


$K \leftarrow \{\}$ \hfill \comment{0.9}{// Set of Gaussians}

\For{$c_i \in c$}{

	$xdist \leftarrow c_i.x -c_{(i+1)mod|c|}.x$
	
	$ydist \leftarrow c_i.y -c_{(i+1)mod|c|}.y$
	
	\If{$ydist < 0$}{ 
	
		$xdist \leftarrow -(xdist)$ \hfill \comment{0.9}{// keep angle between $0^\circ$ and $180^\circ$}
		
		$ydist \leftarrow -(ydist)$
	}

	$\alpha \leftarrow atan2(ydist,xdist)\frac{360}{2\pi}$
	\hfill \comment{0.9}{// angle of contour segment}

\definecolor{light-gray}{gray}{0.95}
\hspace*{-\fboxsep}\colorbox{light-gray}{\parbox{0.95\linewidth}{%

	$\sigma \leftarrow 1$ 
	\hfill \comment{0.9}{// stdev of Gaussian distribution}

	$\lambda \leftarrow \frac{|c_i|}{\sum_{c_i \in c} |c_i|}$ 
	\hfill \comment{0.9}{// normalization factor}
}}

	$f_{c_i, c_{(i+1)mod|c|}}(\theta) = \lambda \frac{1}{\sigma\sqrt{2\pi}} 
	e^{-\frac{1}{2}\left(\frac{\theta-angle(c_i)}{\sigma}\right)^2 }$

	$K \leftarrow K \cup \{ f_{c_i, c_{(i+1)mod|c|}}(\theta) \}$

}

$c_{angle} \leftarrow max_{\theta}(\Sigma_{f \in K} f(\theta)) $

$c_{rot} \leftarrow rotate(c, c_{angle})$

\end{algorithm}

\subsection{Building Candidate Feature Construction}
\label{section:features}

To build a classification model that can filter candidates into building and non-building, we need features that can discriminate effectively and are efficiently computed. Haar features have been shown to quickly capture discriminative contrast patterns effectively \cite{viola_robust_2004}. They are generated by taking a rectangular image mask and dividing it into two rectangles, with a horizontal or vertical division. The sum of the pixel values in one rectangle are subtracted from the sum of the pixel values in the other. Haar features are discriminative in face and crater detection \cite{cohen_crater_2013} because these domains have similar contrast at specific positions of the candidates. In this work each candidate is scaled to 200 x 200 pixels before Haar features are extracted. Horizontal and vertical Haar features are extracted in a sliding window fashion which extracts square regions from the image systematically from the top left to the bottom right. Square regions are extracted with pixel width 40, 80, and 100 are applied with a step size of 10 pixels. Also, square regions are extracted with width 20 with a step size of 5 pixels in order to capture small details. This yields a total of 3592 features. Each feature represents the horizontal or vertical contrast in that region with a signed integer value. A value of 0 means no contrast where a positive or negative value represents contrast in the positive or negative direction. The sign of the number is dependent on the order of the subtraction during extraction and is only useful for comparison.

\begin{figure}[th]
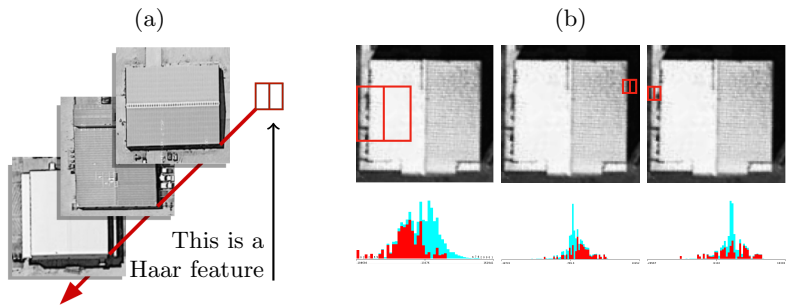

\begin{center}
\centering
\subfloat[]
{%
\begin{tikzpicture}[scale=1]
\node[inner sep=0pt] at (0,0){\includegraphics[width=0.3\columnwidth] {haarexplain.png}};
\draw[thick,->,style={}] (1.7,-1.4) -- node[align=right,left,yshift=-20pt] {This is a\\Haar feature} (1.7,0.7);
\end{tikzpicture}
}%
\hspace{20pt}
\subfloat[]
{%
\begin{tabular}{c}
  \includegraphics[width=0.15\columnwidth]{haar-b-3896.png}
  \includegraphics[width=0.15\columnwidth]{haar-b-1438.png}
  \includegraphics[width=0.15\columnwidth]{haar-b-216.png}
\\
  \includegraphics[width=0.15\columnwidth]{haar-3896-plot.png}
  \includegraphics[width=0.15\columnwidth]{haar-1438-plot.png}
  \includegraphics[width=0.15\columnwidth]{haar-216-plot.png}
\\
\end{tabular}%
}
  
  \caption{(a) Example of a Haar feature being extracted from building candidates at the same position on multiple candidates in order capture the contrast at the edge of the building. (b) The three highest weighted Haar features of a Linear AdaBoost classifier in descending from left to right. The distribution of values extracted from Dataset A for each feature is shown at the bottom to show their linear separability. }
  \label{fig:haar}
\end{center}
\end{figure}

By aligning buildings and adding padding to expose its edges, which have high contrast, we are able to obtain contrast patterns between candidates. For example the Haar features being extracted in Figure \ref{fig:haar}a will statistically expose higher contrast in candidates which contain buildings due to the edges of this buildings appearing in the same location across examples. Also, roof texture and the surrounding area texture may also be consistent enough to provide linear separable distributions of values with respect to a building and non-building. In order to gain more insight we analyze the top weighted Haar features in the Linear AdaBoost classifier in Figure \ref{fig:haar}b where it can be seen that edges of buildings are very discriminative. We are able to conclude that the statements from previous work that find shadows a dominant feature are correct. Shadows will generally exist at the edges of buildings and provide strong contrast values at the edge of the roof where the shadow begins. Together, many of these features allow us to obtain a linear separable feature space to achieve accurate classification. One problem that arises from using these features is when buildings have black roofs the contrast between the roof and the shadow is very low and might appear to be very similar to a solid surface.

\subsection{Candidate Permutation Search (PHM)}
\label{section:phm}

\begin{figure}[h]
\begin{center}
  \includegraphics[width=0.427\columnwidth]{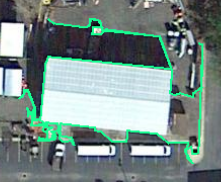}
  \includegraphics[width=0.45\columnwidth]{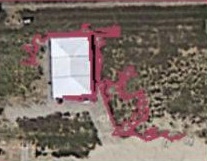}
  \caption{Example of contours that overdetected a candidate. The green and red lines are the contour lines. The bounding boxes can be repositioned to detect these buildings.}
  \label{fig:missalignment}
\end{center}
\end{figure}

Some candidates are lost during the initial preprocessing step due to contours that cover part (or too much) of the building as shown in Figure \ref{fig:missalignment} This leads to a misalignment of Haar features.

\begin{figure}[h]
\begin{center}
  \includegraphics[width=0.9\columnwidth]{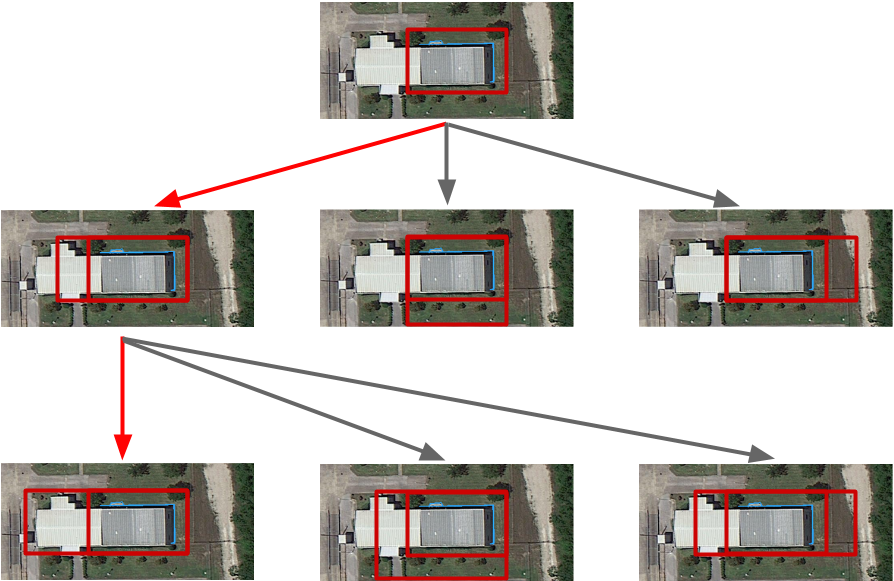}
  \caption{An example of the PHM search space being traversed in a greedy manner. Each potential permutation becomes a link which represents a new frame that Haar features are extracted from. The red lines indicate the path taken during the search to cover candidates.
  }
  \label{fig:phmsearch}
\end{center}
\end{figure}

To solve this problem we present a Permutable Haar Mesh (PHM) algorithm which iteratively permutes the building candidate using a custom heuristic function to search the space shown in Figure \ref{fig:phmsearch}.  We perform a multi objective greedy search (for speed) using the following function (for accuracy) based on the result of a classifier:
$$\mathbb{H}(can,{\cal L}) = 
\frac{2\overbrace{{\cal L}^+ (can)}^{\text{P(bldg)}}
(1-\overbrace{{\cal L} ^- (can))}^{\text{P(not bldg)}}} {{\cal L}^+(can) +
(1-{\cal L}^- (can))} $$
Here we take the harmonic mean of ${\cal L}^+ (can)$, the probability that $can$ is a building, and $(1-{\cal L} ^- (can))$, the complement of the probability that $can$ is not a building. Using a greedy search we evaluate each permutation and select the best increase in probability at each step of the iteration until we cannot improve the hypothesis probability.  This method is outlined in Algorithm \ref{alg:permute}.

\begin{algorithm}[!h]
\caption{Greedy Candidate Permutation Search}
\label{alg:permute}
\KwIn{Candidate $can$\\
\hspace{28pt} Permutation Rate $r$\\
\hspace{28pt} Heuristic function $\mathbb{H}$}

\KwOut{Best Candidate $best$\\}

\vspace{5pt}

$d = r\cdot \frac{(bottom-top)+(right-left)}{2}$
\label{alg:permute:distancefromrate}

$P \leftarrow \{ can+(d \cdot p) | p \in \{ 
\left(
\begin{array}{c}
x_0\\
x_1\\
x_2\\
x_3
\end{array}
\right) |
x_i \in \{0, -1, 1\}
\}$

$
best \leftarrow argmax_p(\mathbb{H}(p)) $ where $p \in P
$

\If{$best \neq can$}{
	return $permute(best)$ \hfill \comment{0.9}{//if $\mathbb{H}$ improved continue search}
	}
\Else{return $best$} 
\end{algorithm}

\subsection{Complexity}
\label{section:complexity}

Our method is $O(n)$ for generating candidates which places the training complexity on the classifier used. Each candidate generated as a negative example adds to the complexity. This can be reduced by generating less negative examples but this may also generate a classifier with lower performance.

When utilizing the classifier our method is $O(n)$ in terms of pixels or candidates. In the worst case every pixel could be considered a candidate which would be determined in linear time using Canny edge detection and Chang's linear contour detection, we call this $n$. When sampling and merging using a specific step we incur a fixed cost dependent on the step size chosen. For $0.05$ this is $400$ leading to a potential $400n$ candidates to evaluate. Our rotation method is based on the number of vectors in the contour ($c$) of the candidate. The maximum number of contours would be the number of pixels in the candidate. Our approximation method solves this in $360|c|$. Each candidate then has a fixed number $(4,240)$ of Haar features extracted which is one initial cost of the candidates pixels for an integral image and then 4 additions per Haar feature. When using a linear classification model, such as Naive Bayes or AdaBoost on linear decision stump classifiers, each candidate can then be classified in linear time.

\section{Experimental Evaluation}
\label{section:eval}

In order to evaluate our method we looked for publicly available datasets that would allow us to study the errors when applying methods to the average residential buildings as well as unique industrial buildings. \cite{mnih_learning_2010} has generated a benchmark dataset using MassGIS which contains average residential buildings but industrial buildings such as coal and nuclear power plants are not released by MassGIS. Because of this we have built a dataset of nuclear power plant buildings that can be shared with the research community. We utilize these two datasets in order to showcase the robustness of our algorithm on imagery with various quality and content.

\begin{figure*}[!t]
\begin{center}

\subfloat[Dataset A Ground Truth]
{
\label{fig:datasetagt}
\fcolorbox{gray}{white}{
\includegraphics[width=0.40\textwidth,clip]{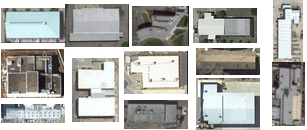}
}}
\hspace{20pt}
\subfloat[Dataset B Ground Truth]
{
\label{fig:datasetbgt}
\fcolorbox{gray}{white}{
{\includegraphics[width=0.40\textwidth]{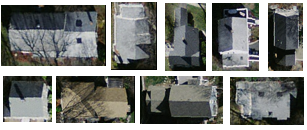}}
}}

\subfloat[Dataset A Negative Examples]
{
\label{fig:datasetatn}
\fcolorbox{gray}{white}{
{\includegraphics[width=0.40\textwidth,clip]{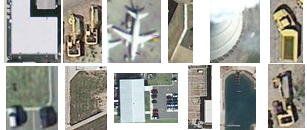}}
}}
\hspace{20pt}
\subfloat[Dataset B Negative Examples]
{
\label{fig:datasetbtn}
\fcolorbox{gray}{white}{
{\includegraphics[width=0.40\textwidth,clip]{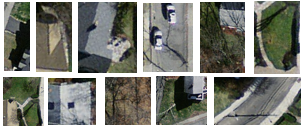}}
}}

\caption{
 Shown here are samples images of the two datasets used in analysis. All images are automatically cropped and rotated based on their contours. At the top we have ground truth buildings and at the bottom are negative examples.}
\label{fig:datasets}

\end{center}
\end{figure*}

\textbf{Dataset A} (Figure \ref{fig:datasetagt}) was constructed using images from Google Maps \footnote{https://maps.google.com/} with various resolution, size, illumination, geographic region, building size, and building purpose. There are 411 buildings in this dataset which are mostly non-residential including large industrial and power generation. These buildings can be very unique to a specific purpose and vary greatly in size.

\textbf{Dataset B} (Figure \ref{fig:datasetbgt}) is a labelled subset of the dataset used in \newline \cite{mnih_learning_2010}\footnote{http://www.cs.toronto.edu/$\sim$vmnih/data/}. We used a higher resolution (15cm/pixel) version of the same imagery acquired from MassGIS (coq2008\_15cm\_jp2). All buildings have the same illumination. This dataset is of a contiguous area composed of mostly residential buildings. In total there are 1337 buildings.

We use these datasets to first evaluate the recall obtained by our method. Recall is an evaluation metric representing how many buildings have not been missed, this metric is complementary to precision which represents how correct each prediction is. After this we discuss how our positive and negative examples are constructed to train a classifier. This is followed by an analysis of candidate alignments effect on these examples on various classifiers. We then discuss how we can increase recall with our PHM method can recover candidates and achieve better accuracy at the cost of a more computationally expensive method. Finally we evaluate the runtime of different components of our algorithm.

\subsection{Candidate Recall}
\label{section:candidates:eval}

\begin{figure}[h]
\begin{center}
\includegraphics[width=0.7\columnwidth,trim = 0mm 0mm 5mm 23mm,clip,
]{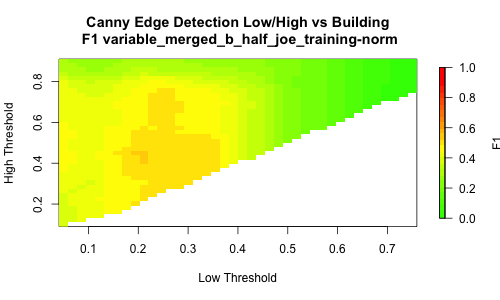}
\caption{Here all possible high and low threshold values from 0,0 to 1,1 for the Canny edge detector are evaluated on dataset A with step size 0.05. The recall value is plotted and we can observe a spike at 0.2,0.4. Further inspection reveals that different buildings are being captures at different combinations resulting is no one maximizing combination of threshold values.}
\label{fig:lowhigh:variablemergedab}
\end{center}
\end{figure}

It is important that we achieve high recall in order to not miss any potential buildings using our candidate generation method. Unfortunately there are some complications that we had to overcome.  Using a single high and low Canny threshold value we are only able to achieve low recall values. In Figure \ref{fig:lowhigh:variablemergedab} we explore all possible configurations of low and high threshold values on dataset A. These results show a strange surface due to a trade off of capturing different sizes of the buildings. There does seem to be a peak but it is very low $\approx 60\%$. Some buildings are only identified as candidates at specific threshold values so changing them misses some while finding others. The problem is that these values are not the same for every building in a dataset as shown in Figure \ref{fig:cannythreshimage}. This observation leads us to our solution, because some buildings are only captured by different threshold values.

\begin{figure}[h]
\begin{center}

\includegraphics[width=0.7\columnwidth]{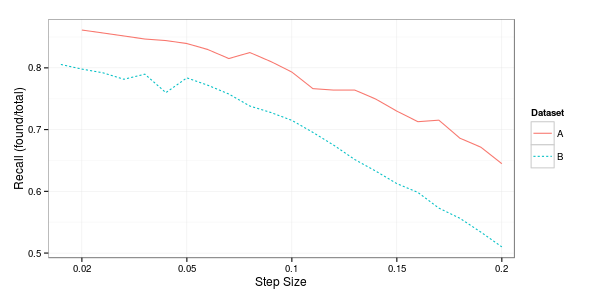}

\caption{We vary the step size used to generate candidates. As we decrease the step size, meaning more samples, the recall increases and we are able to capture more of the buildings.}
\label{fig:lowhigh:recall}
\end{center}
\end{figure}

\begin{figure}[h]
\begin{center}

\includegraphics[width=0.7\columnwidth]{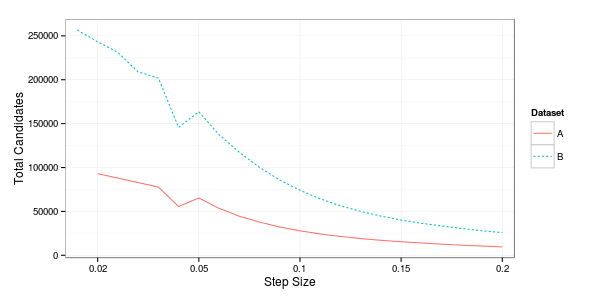}

\caption{We vary the step size used to generate candidates. As we decrease the step size in order to gain higher recall the number of candidates increases.}
\label{fig:lowhigh:total}
\end{center}
\end{figure}

To solve this problem we generate candidates by \textbf{sampling and merging} the results of candidate generation at many different threshold values. The question now is what Low/High threshold values to use. We experiment with various step sizes through the space (0,0) to (1,1) in Figure \ref{fig:lowhigh:recall}. As the step size is reduced from 0.2 to 0.01 the recall increases at a diminishing rate. However, there is trade-off that must be made when choosing a small step size. In Figure \ref{fig:lowhigh:total} the total number of candidates that must be evaluated is analyzed. As the step size is reduced the total number of candidates increases to numbers that are much larger than the number of buildings that exist in those images. This may not only increase running time but also decrease overall performance by increasing the chance that a classifier may misclassify.

To put more context on Figure \ref{fig:lowhigh:recall}, in dataset A we start with 411 labeled buildings and our preprocessing step is able to find 86\% when generating about 90,000 candidates. In dataset B we start with 1,337 labeled buildings and our preprocessing step is able to find 80\% when generating about 240,000 candidates. To put this in perspective, without this preprocessing step, because the centers can be any pixel, the width and height can be any combination (non-overlapping), and the rotation can be between $0^\circ-180^\circ$, a small 1,000 x 1,000 image can easily generate over 1 billion candidates using a sliding window for just one image in order to achieve 100\% recall.

\subsection{Training Set Construction}

To learn an accurate classifier requires constructing a training set containing difficult realistic examples of what will be presented to the classifier during testing. We run the candidate generation process and subtract the positive examples. This process includes candidates that partially overlap the ground truth in order to train on examples that may be misclassified during testing. Our goal is to select strong representative examples that we expect to reside near the decision boundary of a classifier. 

For all the evaluations following this section, 10-fold cross validation is used to calculate the F1-Score obtainable with a classifier. We define the F1-Score as follows:
$$F1 = \frac{2}{\frac{1}{\textit{recall}} + \frac{1}{\textit{precision}}}$$
$$\textit{precision} =\frac{\textit{true positives}}{\textit{true positives}
+ \textit{false positives}}$$
$$\textit{recall} =\frac{\textit{true positives}}{\textit{true positives} + \textit{false
negatives}}$$

Dataset A has 383 positive and 4,912 negative examples. Dataset B has 992 positive and 11,488 negative examples. The number of positive examples is less than the total ground truth number because some candidates are excluded because the 5\% padding that is added goes out of the image bounds and is not included. The datasets are balanced in order for the classifiers to properly learn. This is done by randomly sampling with replacement to add duplicates to the positive examples. 

All experiments are performed with the AdaBoost classifier unless otherwise noted. In the next section we compare many different classifiers. The Weka implementations of these algorithms are used with their default values. 

\begin{itemize}
  \item \textbf{AdaBoost} is an ensemble of weighted linear classifiers with one feature each. The classifier is trained for 10 epochs with a weight threshold of 100 to prune the weights\cite{freund_experiments_1996}.
  \item \textbf{Naive Bayes} assumes all variables are conditionally independent with respect to the class label. This classifier then simply uses Bayes' rule to determine the probability of a class attribute given feature values \cite{john_estimating_1995}.
  \item \textbf{J48} constructs a decision tree by iteratively splitting each tree node if classification error is reduced when discriminated by an attribute. The Weka version is based on the C4.5 implementation by Quinlan and uses the default confidence value of 0.25 to estimate error \cite{quinlan_c4.5:_1993}.
  \item \textbf{Random Forest} constructs decision trees from subsets of features which are drawn uniformly with replacement from the global feature set. 100 trees are constructed. Each decision tree is constructed similar to J48. The resulting classification is a majority vote by all trees for a class label \cite{breiman_random_2001}.
  \item \textbf{Support Vector Machine}: The Weka LibSVM implementation of C-SVC was used as described by \cite{cortes_support-vector_1995}. A radial basis kernel was used with the parameters $\nu = 0.5, \gamma = 0, loss = 0.1, cost = 1$.
\end{itemize}

\subsection{Rotation impact on classifiers}

\begin{figure}[h]
\begin{center}
\includegraphics[width=1\columnwidth]{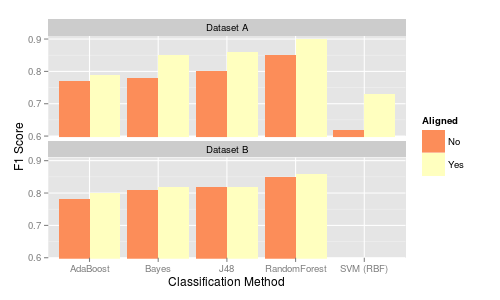}

\caption{We compare AdaBoost (with linear decision stump classifiers), Naive Bayes, J48 Decision Trees, Random Forest, and SVM (with a radial basis function kernel) classifiers applied to datasets A and B via their F1-Score with and without rotation of the candidates.}
\label{fig:classifierrotation}
\end{center}
\end{figure}

Analysis is performed to evaluate the effect of rotating candidates on the overall pipeline. To demonstrate the versatility of this step we evaluate many classifiers. In Figure \ref{fig:classifierrotation} it can be observed that rotating candidates increases the F1-Score of standard classification algorithms.

To evaluate the following classification methods we generate candidates from each training set using the sample and merging method with step size 0.05 and form an isolated set of candidate images so that 10-fold cross validation can easily be performed. The results here are the metrics from these isolated sets and therefore don't reflect the impact of recall loss from the preprocessing method which is analyzed in Section \ref{section:candidates:eval}. 

We evaluate AdaBoost because it was used as part of the Viola and Jones face detection pipeline \cite{viola_robust_2004}. AdaBoost is expected to be well suited for this task because it performs feature selection on the many Haar features generated from the candidate in both situations. This is however not the case. AdaBoost ranks among the worst classifiers evaluated.

We evaluate Naive Bayes and J48 Decision Tree classification models as baselines which are quick to train that we expect the reader will be familiar with. A random classifier was used to confirm 50\% F1-Score indicating balanced training data. We also evaluate Random Forest and find it to outperform all other methods.

The previous classification models discussed can rapidly be trained and utilized in comparison to a Support Vector Machine (SVM) with a non-linear kernel. We were able to evaluate Dataset A using an SVM with a radial basis function kernel. However, due to the computational cost we are unable to evaluate Dataset B using an SVM because the algorithm did not terminate in 72 hours. It is interesting how poorly the SVM model performs. We can speculate that it may be caused by noisy or irrelevant Haar features. A large amount of features may cause the classifier to weight features inappropriately and skew classification. The increase in performance after candidate rotation may indicate this as it causes features to have a higher discriminative ability which can more easily be separated.

Overall, every classification method had its F1-Score increase after the alignment of candidates. The most significant increase was for an SVM classifier.

\subsection{Best PHM Permutation Rate}

\begin{figure}[h]
\begin{center}

\includegraphics[width=1\columnwidth]{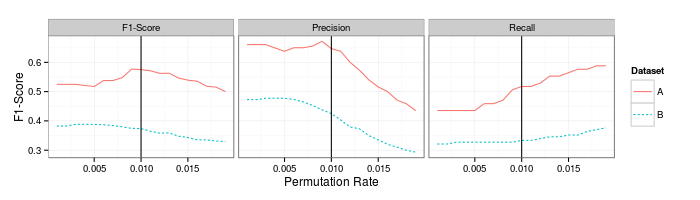}
	\caption{Our pipeline using Canny threshold values of low:0.2/high:0.4 varying the permutation rate on both datasets. A permutation rate of 0.01 is able to increase recall while maintaining precision to yield a higher F1 value.}
	\label{fig:optimalpermutestep}
\end{center}
\end{figure}

The primary goal of our preprocessing method is to maintain high recall. If candidates are still missed we can use the PHM method to salvage over/underdetected candidates as outlined in Section \ref{section:phm}. This method is analyzed in Figure \ref{fig:optimalpermutestep} to study how the F1-Score is impacted as the permutation rate is increased. For these experiments we used one combination of high and low Canny threshold values instead of merging many values together which yields lower recall values from the start.

In Figure \ref{fig:optimalpermutestep} as the rate of permutation increases so does the recall. However, similarly as the permutation rate increases the precision falls. The increase in precision error is due to more candidates being presented to the classifier which appear to be buildings as a result of the PHM process itself. A compromise is found at the peak of the F1-Score plot of 0.01.

\subsection{Linear Time Feature Extraction}
\label{section:lineareval}

Our machine learning pipeline runs in linear time as theoretically explained in Section \ref{section:complexity}. We empirically evaluate the runtime on a single 3.07GHz Intel Xeon CPU. However many parts of the algorithm are easily made parallel to achieve major speed improvements. 

\begin{figure}[ht!]
\begin{center}
 \subfloat[Contour generation]{%
  \includegraphics[width=0.8\columnwidth]{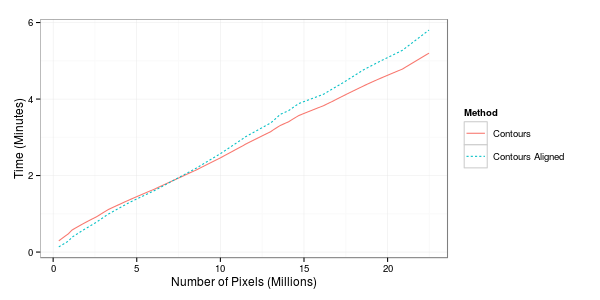} 
  \label{fig:runtime:contours}
 }%
 
 \subfloat[Haar feature extraction]{%
  \includegraphics[width=0.8\columnwidth]{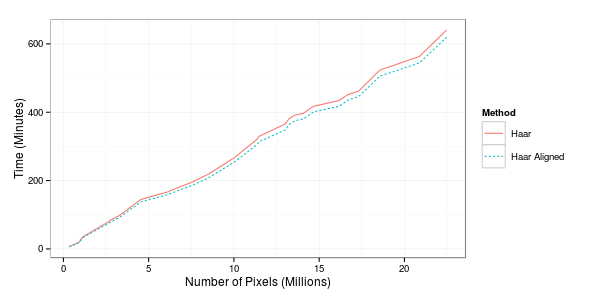} 
  \label{fig:runtime:haar}
 }%
 
\caption{Here we show the impact of rotation on runtime during the contour generation (\ref{fig:runtime:contours}) and Haar feature extraction (\ref{fig:runtime:haar}) parts of the process.}
  
\end{center}
\end{figure}

The first way to empirically show this is during the initial contour extraction phase analyzed in Figure \ref{fig:runtime:contours}. Here images are processed one after another, the total number of pixels processed is plotted against the time taken. Here it is observed that aligning the contours only slightly increases the processing cost. 

In Figure \ref{fig:runtime:haar} we perform the same evaluation but allow the process to continue to the step of extracting Haar features from every candidate. A strange result is that it takes less time when we add the rotation step. An answer for this may be that the scaling phase before Haar features are extracted is sped up because images contain less edges on diagonals.

\begin{figure}[h]
\begin{center}
  \includegraphics[width=0.9\columnwidth,trim = 0mm 0mm 10mm
5mm,clip]{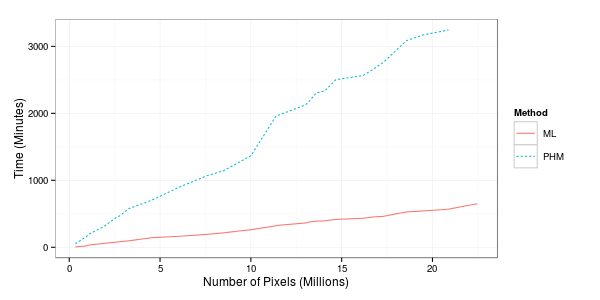}
  \caption{Here the runtime is evaluated using the complete pipeline for our ML and PHM methods.}
  \label{fig:runtime:mlphm}
\end{center}
\end{figure}

In Figure \ref{fig:runtime:mlphm} we evaluate the entire pipeline and observe that our basic machine learning (ML) approach appears significantly faster than PHM. For every candidate encountered during the algorithm the PHM will search possibly 100's of surrounding candidates to find a better match. From our experience the machine learning approach appears to work in almost realtime on reasonably sized images.

\section{Related Work}
\label{section:rw}

Automated labeling of aerial images has been a motivating problem for researchers for a very long time \cite{irvin_methods_1989}. The development of an automated system to identify discrete objects, such as buildings, has been a much sought after goal. Many techniques from the field of computer vision have been employed, as well and statistical machine learning approaches. A number of surveys including \cite{mayer_automatic_1999}, \cite{baltsavias_object_2004}, and \cite{druaguct_automated_2006} indicate the depth of this field.

Unlike our method which relies only on RGB images, much work has been done using very high spatial resolution (VHR) multispectral data, \cite{sohn_data_2007} synthetic aperture radar (SAR) data \cite{simonetto_rectangular_2005} and light detection and ranging (LIDAR). This information has been used to filter out sections of images corresponding to non-building areas such as vegetation or water. Information such as azimuth and zenith angles has been used to calculate the shadow locations and near infrared to better determine building shadows from plant shadows \cite{ok_automated_2013}.

Working only with images, other researchers have explored techniques using many different types of features that can capture texture information, color, shape, and contextual information. Simple features can be built using the color and intensity of pixels, and gradient based features have also been used. Local scale and rotation invariant features like Lowe's SIFT \cite{lowe_distinctive_2004} and the sped up version SURF\cite{bay_surf:_2006} have been evaluated \cite{yang_geographic_2013} \cite{sirmacek_probabilistic_2011}.

Shadows have been picked up as a powerful building indicator that can be identified by simple algorithms similar to ours \cite{irvin_methods_1989} and \cite{wei_urban_2004}. Machine learning has been employed extensively, with various systems using features to train classifiers such as Support Vector Machines \cite{mountrakis_support_2011}. Lately, deep learning techniques such as Convolutional Neural Networks have been used to good effect \cite{mnih_learning_2010}.

Our method stands out from these other approaches because of our focus on speed and applicability to all geospatial imagery because our method only needs pure RGB images and does not require a near-infrared channel or azimuth and zenith angles. Also, unlike other methods we provide an implementation of our method.

\section{Conclusion}
\begin{figure*}[htp]
\begin{center}
  \includegraphics[width=0.5\columnwidth]{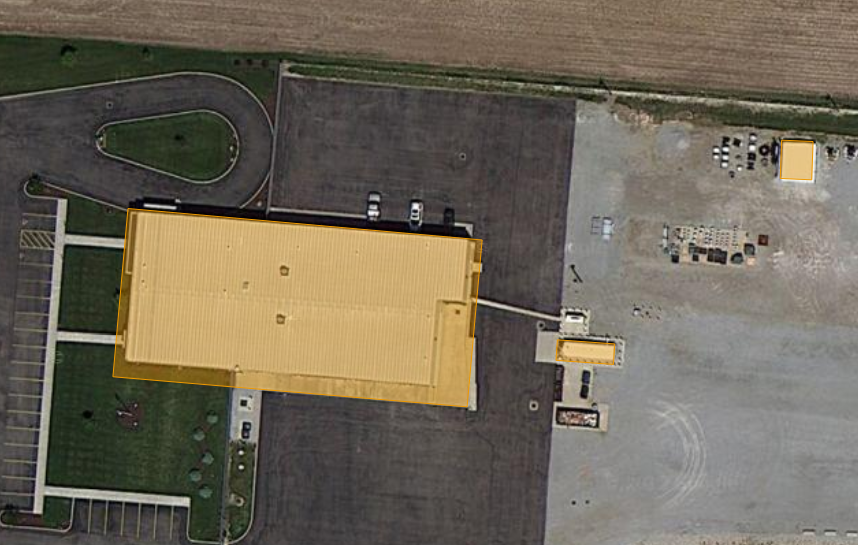}%
  \vspace{1pt}%
  \includegraphics[width=0.5\columnwidth]{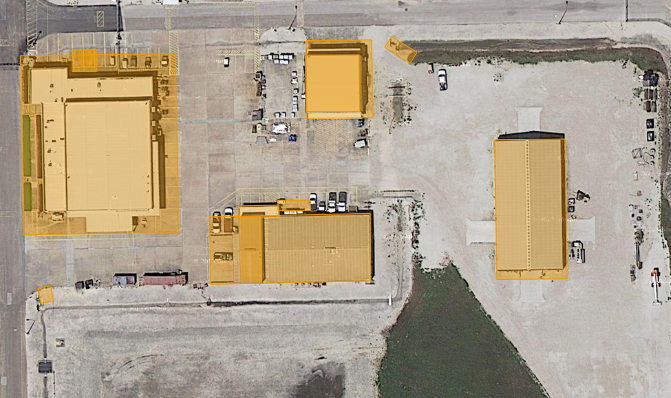}%
  \vspace{1pt}
  \includegraphics[width=0.5\columnwidth]{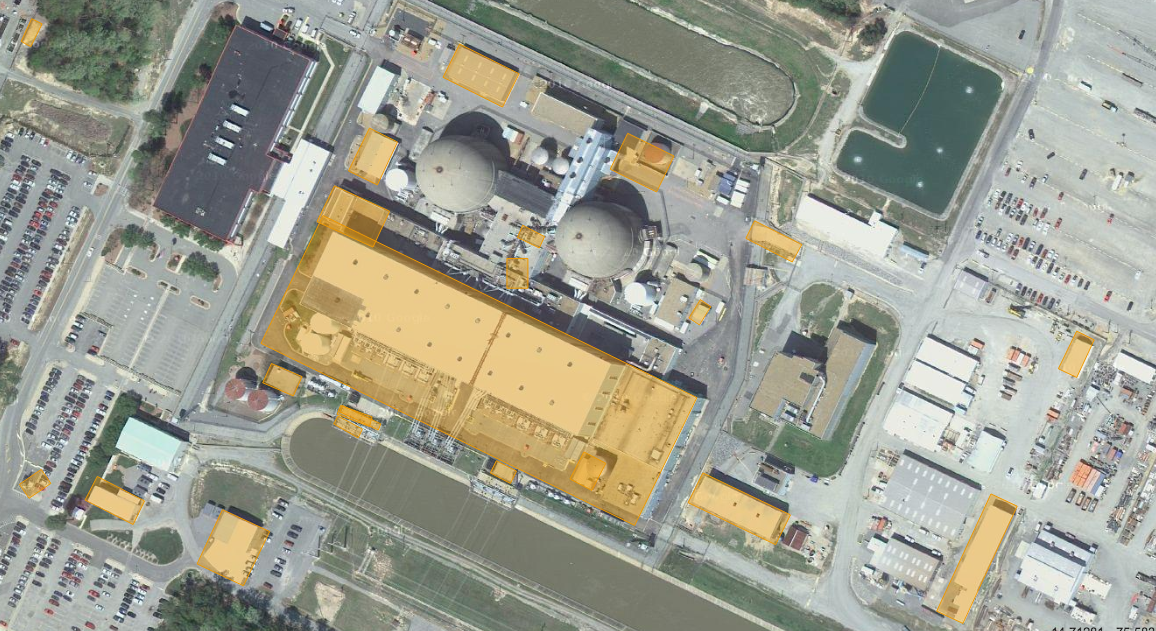}%
    \vspace{1pt}%
  \includegraphics[width=0.5\columnwidth]{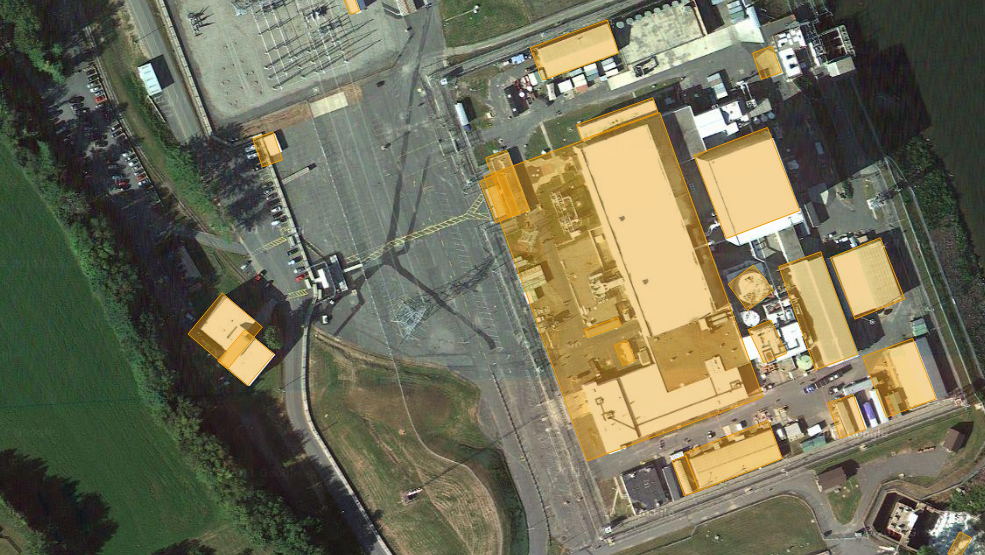}%
  \vspace{1pt}
  \includegraphics[width=0.5\columnwidth]{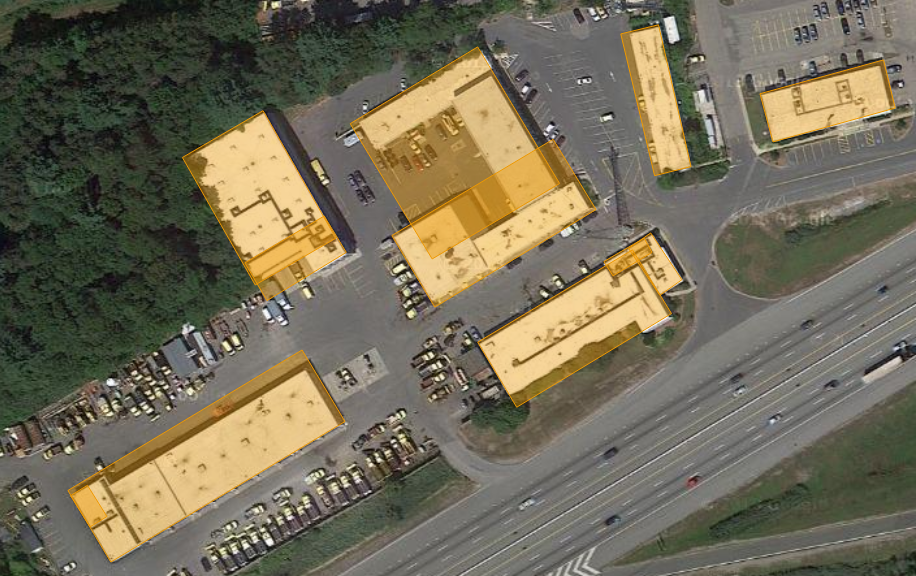}%
  \vspace{1pt}%
  \includegraphics[width=0.5\columnwidth]{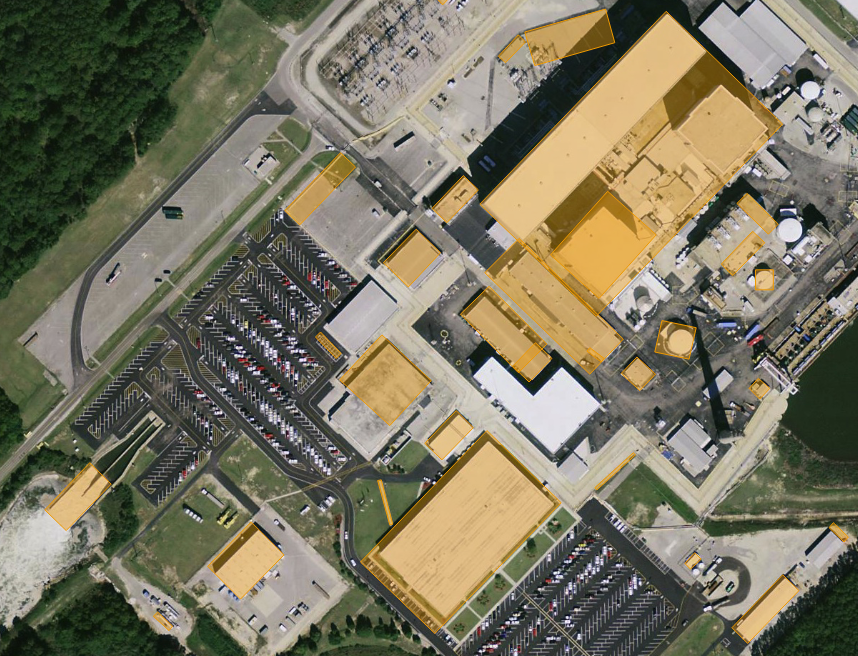}%
  \vspace{1pt}
  \includegraphics[width=0.8\columnwidth]{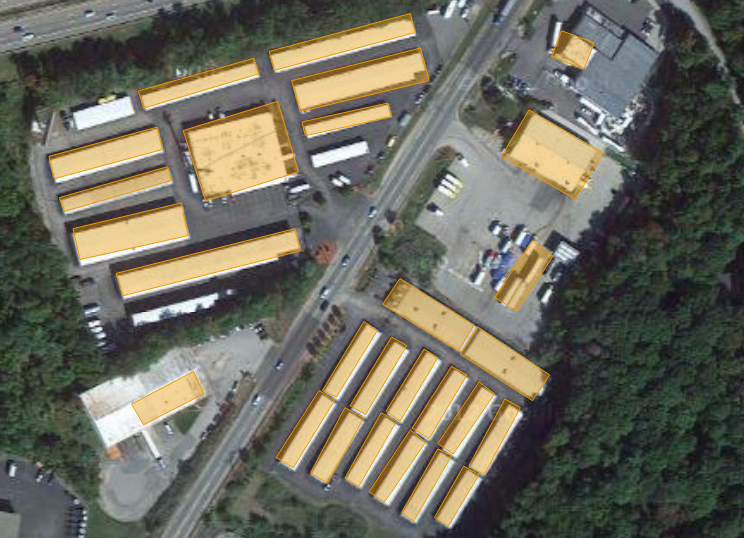}
  \caption{Some images from Dataset A are analyzed with our machine learning method using an AdaBoost classifier. Predictions are highlighted in yellow. On these examples we detect over 90\% of the buildings except on heavily clustered buildings around nuclear power plants which present a difficult task because candidate building borders abut each other and prevent shadows.}
  \label{fig:showcase}
\end{center}
\end{figure*}

In this paper we describe algorithms for reducing discrete object detection in reflective geospatial imagery to machine learning, specifically in the case of buildings. Results from the application of this method are shown in Figure \ref{fig:showcase}. We have shown the complex patterns of a discrete object's contrast features can be learned using state of the art ML methods. The reduction requires non-trivial ML-aware preprocessing methods. We have shown that these methods consistently increase the performance of classification algorithms. We also present the concept of a PHM in order to recover candidates that fail to be classified correctly. This method generates a search space which has potential to greatly increase detection rates and requires further research to fully utilize beyond what is explored in this paper.

\section{Acknowledgements}
This work is partially funded by a grant from the National Nuclear Security Agency of the U.S. Department of Energy (grant number: DE-NA0001123) as well as by the National Science Foundation Graduate Research Fellowship Program (grant number: DGE-1356104). This work utilized the supercomputing facilities managed by the Research Computing Department at the University of Massachusetts Boston as well as the resources provided by the Open Science Grid, which is supported by the National Science Foundation and the U.S. Department of Energy's Office of Science.

\bibliographystyle{apalike}
\bibliography{joe,objectdetect,landforms,continuous} 

\end{document}